\documentclass[12pt]{elsarticle}

\usepackage{amssymb}

\usepackage{subcaption}
\usepackage{amsmath}
\usepackage{multirow}
\usepackage{xcolor} 
\usepackage[nobiblatex]{xurl}

\journal{"Preprint. Under review"}

\begin{document}

\begin{frontmatter}

\title{Spectral Image Data Fusion for Multisource Data Augmentation}

\author{Roberta Iuliana LUCA\footnote{Faculty of Mathematics and Computer Science, \textit{Transilvania} University of Brașov, Romania, e-mail: roberta.luca@student.unitbv.ro}}

\author{Alexandra B\u AICOIANU\footnote{Department of Mathematics and Computer Science, \textit{Transilvania} University of Bra\c sov, Romania, e-mail: a.baicoianu@unitbv.ro}}

\author{Ioana Cristina PLAJER$^{*,}$\footnote{${^*}$ {\it Corresponding author}, Department of Mathematics and Computer Science, \textit{Transilvania} University of Bra\c sov, Romania, e-mail: ioana.plajer@unitbv.ro}}

\begin{abstract}

Multispectral and hyperspectral images are increasingly popular in different research fields, such as remote sensing, astronomical imaging, or precision agriculture. However, the amount of free data available to perform machine learning tasks is relatively small. Moreover, artificial intelligence models developed in the area of spectral imaging require input images with a fixed spectral signature, expecting the data to have the same number of spectral bands or the same spectral resolution. This requirement significantly reduces the number of usable sources that can be used for a given model. The scope of this study is to introduce a methodology for spectral image data fusion, in order to allow machine learning models to be trained and/or used on data from a larger number of sources, thus providing better generalization. For this purpose, we propose different interpolation techniques, in order to make multisource spectral data compatible with each other. The interpolation outcomes are evaluated through various approaches. This includes direct assessments using surface plots and metrics such as a Custom Mean Squared Error (CMSE) and the Normalized Difference Vegetation Index (NDVI). Additionally, indirect evaluation is done by estimating their impact on machine learning model training, particularly for semantic segmentation.
\end{abstract}


\end{frontmatter}


\section{Introduction}
\label{sec:introduction}

The ongoing progress in remote sensing sensors helps us to better understand various phenomena around us \cite{msi_fusion}. Each type of material possesses different spectral characteristics that alter how the light is absorbed. This leads to unique spectral fingerprints. Unlike RGB images that only have three spectral bands, multispectral (MS) and hyperspectral (HS) sensors are capable of capturing data from tens or even hundreds of different wavelengths. As a result, each pixel in the image holds a wealth of information, including spectral data related to the chemical composition of the objects \cite{msi_rgb}. These HS sensors can be found on spaceborne platforms like Hyperion \cite{hyperion}, PRISMA \cite{prisma}, or AVIRIS \cite{aviris}, or on aircraft equipped with sensors such as APEX \cite{apex}. The information given by the high spectral resolution sensors is extremely valuable for enabling accurate classifications and detection of pure materials. Therefore, MS and HS images can be very useful in numerous domains like agriculture \cite{msi_app_agriculture}, coastal areas monitoring \cite{msi_app_coastal_monitor}, mineral detection \cite{msi_app_mineral_detection}, and military applications \cite{msi_app_military}.

Although the interest in spectral images is growing and the public availability of more and more satellite data increases there are serious limitations in using them for machine learning tasks, as the different satellites and other equipment produce images with different spectral signatures and spatial resolutions. Additionally, in this context, labeled data for training and testing the models is sparse, which further restricts the datasets that can be practically used. Therefore, we can say that is exceptionally difficult to generalize models for the available data without a preprocessing step. Such a step should enable the fusion of the data based on their characteristics, leading to a uniform and consistent dataset that can be used in any machine learning model. Moreover, it could enable a model trained on a certain dataset to infer spectral data from other sources or sensors. Unfortunately, the research done in this direction is very limited and a standard solution is yet to be found. Moreover, it is crucial to analyze the available data and decide objectively if the fusion is indeed possible for a given task, by example segmentation.

Semantic segmentation of MS and HS images is an important operation in the area of remote sensing. It plays a fundamental role in tasks such as land cover classification, monitoring environmental changes, and overseeing military spaces by labeling each pixel to its respective category. Lately, this task has gained significant attention in the sector of machine learning research. Traditional approaches require human assistance to identify and design the features that will then be used for image segmentation. In contrast, the methods based on machine learning can automatically perform this task in an end-to-end manner by training specific neural networks \cite{msi_semantic_segmentation}, \cite{msi_semantic_segmentation_resnet}, \cite{hsi_semantic_segmentation}. These networks can range from shallow ones, like fully connected neural networks (FCNN) \cite{fcnn}, to more complex and deep ones, often with better results in semantic segmentation, such as convolutional neural networks (CNN)\cite{convolutional_nn}.

In this paper, we introduce a methodology for MS and HS data fusion, aiming to create a larger collection of images that can be used in different machine-learning algorithms. The process of data fusion has always been a complex task and still represents a challenge. It has been extensively researched in machine learning and big data, aiming to optimize and improve the overall data quality. Furthermore, this operation is considered to be an integral component of the applications developed in the present \cite{data_fusion_big_data}, \cite{data_fusion_multisource}.

Moreover, in this study we analyze the available datasets, identify the similarities and differences, choose a reference dataset, and execute an interpolation operation to aggregate all images into a single dataset. Furthermore, to demonstrate this approach's utility, correctness, and accuracy, we also introduce two neural networks specialized in semantic segmentation, which are trained and tested on the aggregated dataset. The networks are also tested on images from other sources, which were not used for the training. Since both FCNNs and CNNs are employed as backbones in the field of spectral image classification and semantic segmentation, we will evaluate our results on both types.

\section{Materials and Methods}
\label{sec:materials_methods}

To perform the data fusion, our initial step involved a comprehensive analysis of the publicly available MS and HS datasets, trying to understand their characteristics and selecting those we considered appropriate for our study. Then, we aggregated the data using various interpolation methods to create a unitary dataset. 

Two key methods were used to validate the suggested methodology for spectral data fusion. Through direct comparison of the interpolated spectral image with the original using some of the metrics outlined in Section \ref{sec:results}, and indirectly through an examination of the segmentation accuracy of the networks trained on the combined datasets.

\subsection{Dataset Description}

The following public MS and HS datasets, with various spectral signatures, were used for the experiments. Firstly we selected Pavia University, Kennedy Space Center (KSC), Botswana, and Indian Pines \cite{hsi_data}, which also come with labels for classification. These were chosen because they share some of the classes, enabling the testing of the interpolation results on classification tasks with neural networks. In order to further analyze the accuracy of the different interpolation methods, we additionally used the well-known CAVE \cite{cave} and UGR \cite{ugr} datasets.

Given the difficulty in finding several MS and HS images with the same labeled classes, we decided to combine the existing labels into two main categories that can be easily found in multiple scenes: \textit{Vegetation} and \textit{Non-Vegetation}. For each of the datasets, an illustration of the original labels and the fused ones is provided in the following.

\subsubsection{Pavia University}

\begin{figure}[htbp]
     \centering
     \begin{subfigure}[]{0.3\textwidth}
         \centering
         \includegraphics[width=0.7\textwidth]{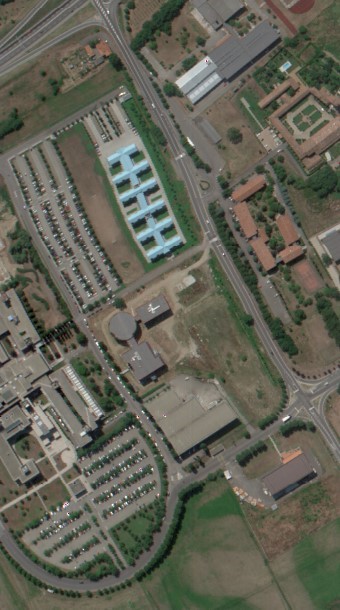}
         \caption{}
     \end{subfigure}
     \hfill
     \begin{subfigure}[]{0.3\textwidth}
         \centering
         \includegraphics[width=0.7\textwidth]{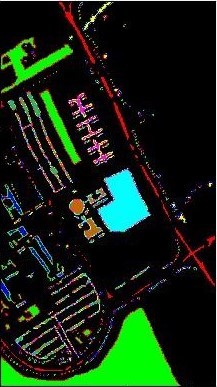}
         \caption{}
     \end{subfigure}
     \hfill
     \begin{subfigure}[]{0.3\textwidth}
         \centering
         \includegraphics[width=0.7\textwidth]{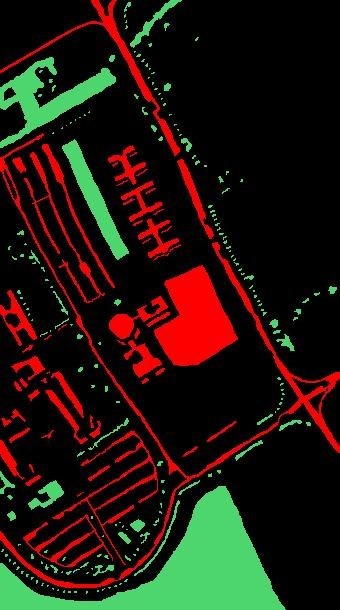}
         \caption{}
     \end{subfigure}
        \caption{Pavia University: (a) Visualization using 3 bands; (b) Original Ground Truth; (c) Processed Ground Truth (\textit{black} for unknown; \textit{green} for vegetation; \textit{red} for non-vegetation)}
        \label{fig:pavia_univ_labels}
\end{figure}
The Pavia University image was obtained in 2001 by the Reflective Optics System Imaging Spectrometer (ROSIS) sensor. It was captured during a flight campaign over Pavia, Northern Italy. The uncorrected data consists of 610 x 610 pixels and 115 spectral bands between 430 and 860 nm, with a spectral resolution of 4 nm. However, some samples did not contain useful information, so they were removed. This resulted in a corrected dataset of 610 x 340 pixels with 103 spectral bands and a spatial resolution of 1.3 m per pixel \cite{hsi_data}. The image is divided into nine ground truth classes. Figure \ref{fig:pavia_univ_labels} illustrates the results for these classes as well as for the two ones considered in our study.

This dataset is frequently explored in the literature for tasks related to image classification using different types of Convolutional Networks. An example of this is the Multiscale Spectral-Spatial Convolutional Neural Network presented in \cite{paviaU_spectral_spatial_cn}.

\subsubsection{Kennedy Space Center (KSC)}

\begin{figure}[htbp]
     \centering
     \begin{subfigure}[b]{0.3\textwidth}
         \centering
         \includegraphics[width=\textwidth]{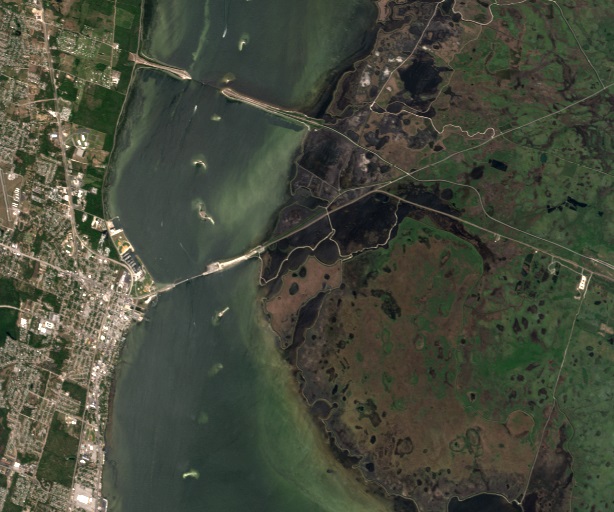}
         \caption{}
     \end{subfigure}
     \hfill
     \begin{subfigure}[b]{0.3\textwidth}
         \centering
         \includegraphics[width=\textwidth]{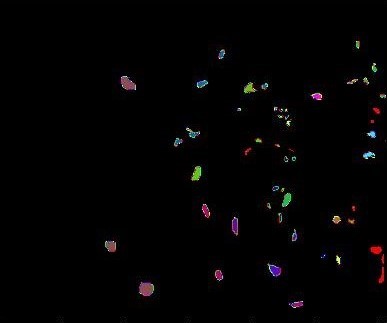}
         \caption{}
     \end{subfigure}
     \hfill
     \begin{subfigure}[b]{0.3\textwidth}
         \centering
         \includegraphics[width=\textwidth]{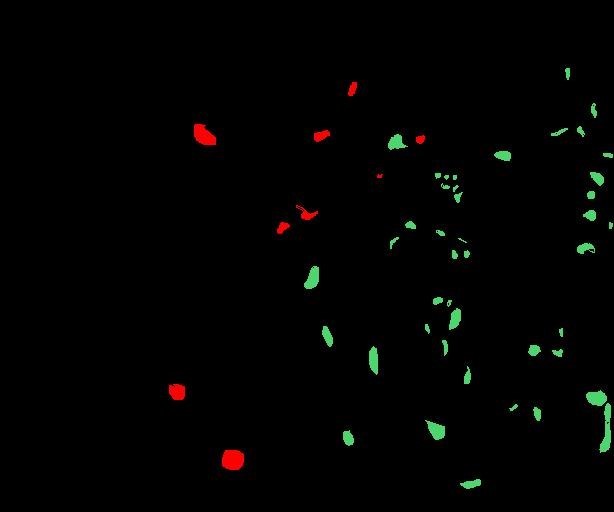}
         \caption{}
     \end{subfigure}
        \caption{KSC image: (a) Visualization using 3 bands; (b) Original Ground Truth; (c) Processed Ground Truth (\textit{black} for unknown; \textit{green} for vegetation; \textit{red} for non-vegetation)}
        \label{fig:KSC_labels}
\end{figure}

The KSC image, captured in 1996 by the NASA Airborne Visible/Infrared Imaging Spectrometer (AVIRIS) sensor, provides a view over the Kennedy Space Center, Florida. It contains a wide wavelength range of 224 spectral bands between 400 and 2500 nm, with a spectral resolution of 10 nm. Among these bands, 48 had to be removed (water absorption and low SNR bands), resulting in 176 bands available for analysis. The image, with a resolution of 512 x 614 and a spatial resolution of 18 m per pixel, originally contained 13 ground truth classes \cite{hsi_data}. The ground truth for these classes, as well, as for the two classes considered by our study, are presented in Figure \ref{fig:KSC_labels}.

KSC is well-known for its applications in classification tasks \cite{hsi_classification} but also in the field of band selection based on different criteria. The study presented in \cite{ksc_band_selection} uses this dataset to demonstrate the accuracy of extracting discriminative properties from HS images.

\subsubsection{Botswana}

\begin{figure}[htbp]
     \centering
     \begin{subfigure}[b]{0.7\textwidth}
         \centering
         \includegraphics[width=\textwidth]{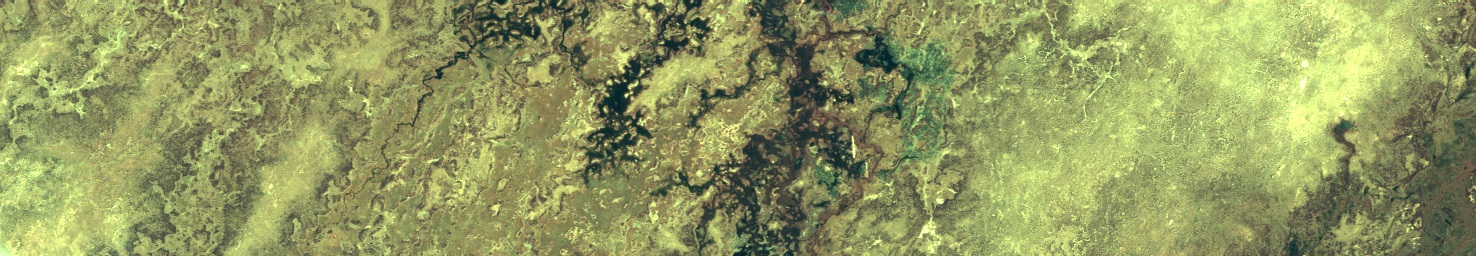}
         \caption{}
     \end{subfigure}
     \hfill
     \begin{subfigure}[b]{0.7\textwidth}
         \centering
         \includegraphics[width=\textwidth]{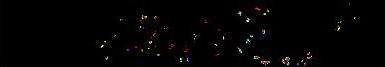}
         \caption{}
     \end{subfigure}
     \hfill
     \begin{subfigure}[b]{0.7\textwidth}
         \centering
         \includegraphics[width=\textwidth]{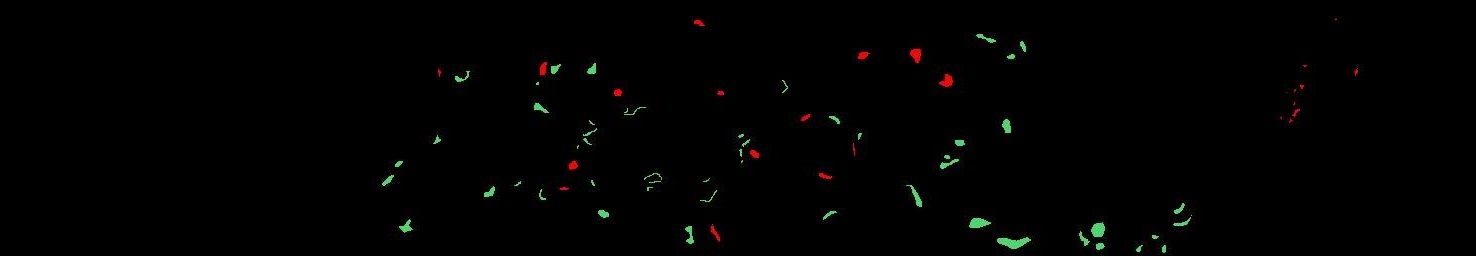}
         \caption{}
     \end{subfigure}
        \caption{Botswana image: (a) Visualization using 3 bands; (b) Original Ground Truth; (c) Processed Ground Truth (\textit{black} for unknown; \textit{green} for vegetation; \textit{red} for non-vegetation)}
        \label{fig:botswana_labels}
\end{figure}

The Botswana image was collected in 2001 by the Hyperion sensor on EO-1 over the Okavango Delta, Botswana. The original uncorrected data consists of 1476 x 256 pixels, each with a spatial resolution of 30 m and 242 spectral bands, spanning from 400 to 2500 nm, with a sampling interval of 10 nm. After the removal of uncalibrated and noisy bands, 145 bands remain for further use \cite{hsi_data}. The original 14 labels and the two merged labels are presented in Figure \ref{fig:botswana_labels}.

Apart from classification tasks, Botswana served as an essential tool for experimental tasks in the field of hyperspectral unmixing \cite{botswana_unmixing}.

\subsubsection{Indian Pines}

\begin{figure}[htbp]
     \centering
     \begin{subfigure}[b]{0.3\textwidth}
         \centering
         \includegraphics[width=\textwidth]{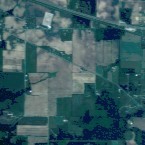}
         \caption{}
     \end{subfigure}
     \hfill
     \begin{subfigure}[b]{0.3\textwidth}
         \centering
         \includegraphics[width=\textwidth]{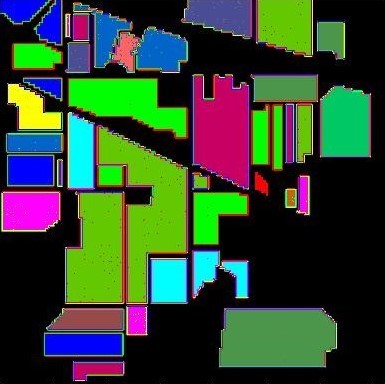}
         \caption{}
     \end{subfigure}
     \hfill
     \begin{subfigure}[b]{0.3\textwidth}
         \centering
         \includegraphics[width=\textwidth]{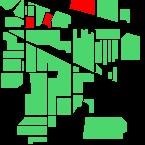}
         \caption{}
     \end{subfigure}
        \caption{Indian Pines image: (a) Visualization using 3 channels; (b) Original Ground Truth; (c) Processed Ground Truth (\textit{black} for unknown; \textit{green} for vegetation; \textit{red} for non-vegetation)}
        \label{fig:indian_pines_labels}
\end{figure}

The Indian Pines dataset was gathered in 1992 by the Airborne Visible/Infrared Imaging Spectrometer (AVIRIS) sensor over the Indian Pines test site in northwestern Indiana. The uncorrected data consists of 145 x 145 pixels, with a spatial resolution of 20 m. The 224 spectral bands range from 400 to 2500 nm, having a spectral resolution of 10 nm. After removing the bands that cover the water absorption region, 200 bands remained in the corrected dataset \cite{hsi_data}. Initially, there were 16 ground truth classes. Figure \ref{fig:indian_pines_labels} shows these classes and the merged two classes considered.

The Indian Pines dataset represents a valuable hyperspectral image used across a broad spectrum of applications. These include classification tasks using Fast 3D CNNs, as presented in \cite{indian_pines_fast_cnn}. It also finds extensive use in the field of super-resolution classification \cite{indian_pines_super_resolution} and band selection studies \cite{indian_pines_bands_selection}.

\subsubsection{CAVE}

The CAVE dataset consists of 32 MS scenes, each of 512 x 512 pixels, captured using a cooled CCD camera. There are 31 spectral bands, spanning from 400 nm to 700 nm, with a spectral resolution of 10 nm. All scenes were captured indoors under controlled lighting conditions. There are no segmentation labels provided. 

This dataset is appropriate for applications in the domain of RGB visualization \cite{msi_rgb}, using various artificial intelligence methodologies. Additionally, it is used for the fusion of multispectral channels \cite{cave_channels_fusion} to improve the image quality.

\subsubsection{UGR}

The UGR dataset contains 14 outdoor urban MS scenes, each captured using a V-EOS HS camera by Photon. The images have a resolution of 1000 x 900 pixels. There are 61 spectral bands between 400 nm and 1000 nm, with a spectral resolution of 10 nm. Similarly, no segmentation labels are provided. 

This multispectral image is used in various scenarios, particularly for the evaluation of machine learning models that are specialized in detecting anomalies, as referenced in \cite{ugr_anormaly_detection}.

\subsection{Interpolation Approach}

In practice, the MS and HS data fusion concept is associated with generating integrated data from multiple MS and HS sources \cite{data_fusion_1}. However, directly merging different datasets into a single one may not produce a usable dataset for a machine-learning task. Traditional neural networks require the input data to be in a consistent format, but gathering a substantial amount of such data for MS and HS images can be challenging. The proposed approach aims to show that a preprocessing step applied to the samples in the datasets, which will then be used in the neural network, can address this issue. This step intends to transform the data to a predetermined format based on the available datasets using the concept of interpolation.

Different interpolation techniques are used in the area of digital image processing \cite{interpolation_image_processing} because they can improve the quality of the output in different scenarios. Interpolation can be used to determine the unknown value of a sample from known values present in a dataset. Moreover, using interpolation for MS and HS images may not only generate values for different channels but may also increase the quantity of the information contained in each pixel.

We needed to establish a reference for the interpolation to properly aggregate or fuse the selected spectral datasets. Most of the chosen datasets have a spectral resolution of 10 nm, except for the Pavia University scene, which has a spectral resolution of 4 nm. As this image, has the best resolution in terms of the wavelength spectrum, it is the most appropriate as a reference for the interpolation. 
In this way, by increasing the number of wavelengths considered for each image, we expect to improve the quantity of the information contained in each pixel. Additionally, given that the maximum wavelength for the CAVE dataset is 700 nm, we decided to further limit the wavelengths to a maximum of 690 nm. This decision was made because interpolating values for the interval [700 - 860 nm] for images in this dataset would have produced unrealistic results. 

Considering Pavia University as the reference dataset, for each wavelength of any other image, the value of each Pavia University channel is interpolated from the values of the adjacent channels of the processed image. We used four different interpolation methods on the five selected images and compared their results. The methods used are well-known in the literature for applications in different domains, including image processing, and will be presented subsequently.

\subsubsection{Linear Interpolation}

Linear interpolation is a mathematical method that generates new data points within the boundaries of a set of known discrete data using linear functions. The formula used to calculate the interpolated value $y$ for a new point $x$ is given by:

\begin{equation}
y = y_1 + (x - x_1) \frac{(y_2-y_1)}{(x_2-x_1)}
\end{equation}
\noindent
where $(x_1,y_1)$ and $(x_2, y_2)$, are the known data points.

This method has been used since antiquity for finding missing values in tables, and it is successfully used in various domains such as medical science \cite{linear_interpolation_med}, simulations \cite{linear_interpolation_simulation}, electronics \cite{linear_interpolation_electronics} and image processing \cite{linear_interpolation_images}.

\subsubsection{Quadratic Interpolation}

Quadratic interpolation assumes that the points follow a parabolic curve modeled by a quadratic equation with the following general form: $y=ax^2+bx+c$. Having three known data points $(x_0, y_0)$, $(x_1, y_1)$ and $(x_2, y_2)$, the following quadratic equation is used to estimate the value of a new point:

\begin{equation}
y = y_0 \times L_0(x) + y_1 \times L_1(x) + y_2 \times L_2(x)
\end{equation}
\noindent
where $L_0(x), L_1(x), L_2(x)$ are the Lagrange basis polynomials. 

This method is widely used in data analysis and curve fitting, but it also has applicability in the image resampling field \cite{quadratic_interpolation_images}.

\subsubsection{Cubic Spline Interpolation}

Cubic spline interpolation, a more complex procedure, is a mathematical method used to create a curve that connects data points with a degree of three, using piecewise third-degree polynomials. It is a special case for Spline interpolation, which successfully avoids Runge's phenomenon \cite{runge_phenomenon}. 

Given $n+1$ data points $(x_i, y_i)$ for $i = 0,1,...,n$, the cubic spline interpolation creates a set of cubic polynomials $C_i(x)$ defined on each interval $[x_i, x_{i+1}]$ as follows:

\begin{equation}
S(x) = \begin{cases} C_1(x), & x_0\leq x \leq x_1 \\ ... \\ C_i(x), & x_{i-1}\leq x \leq x_i \\ ... \\ C_n(x), & x_{n-1}\leq x \leq x_n \end{cases}
\end{equation}

\noindent
where each $C_i = a_i + b_ix + c_ix^2 + d_ix^3$, $d_i \neq 0$, $i = 1,...,n$ is a cubic function.
The coefficients for the cubic spline $S(x)$  $a_i,b_i,c_i,d_i$ are calculated for each interval $[x_i, x_{i+1}]$ by solving the equations:

\begin{equation}
\begin{split}
C_i(x_{i-1}) = y_{i-1}, & i = 1,...,n\\
C_i(x_i) = y_i, & i = 1,...n\\
C^\prime _i(x_i) = C^\prime _{i+1}(x_i), & i = 1,...,n-1\\
C^{\prime \prime}_i(x_i) = C^{\prime \prime}_{i+1}(x_i), & i = 1,...,n-1\\
\end{split}
\end{equation}

This method is commonly employed in computer graphics, animations, robots \cite{cubic_interpolation_applications}, and image magnifying \cite{cubic_interpolation_images}.

\subsubsection{Piecewise Cubic Hermite Interpolating Polynomial}

Piecewise cubic Hermite interpolating polynomial, or PCHIP, is a piecewise polynomial function with a degree of three. It maintains the original shape of the data and, unlike cubic splines, aims to match only the first-order derivatives at the data points with those of the preceding and following intervals \cite{pchip}. We used SciPy from Python \cite{scipy} for the implementation.

\section{Results} 
\label{sec:results}

We used two main approaches to validate the proposed method and compare the results of the interpolations. The first one is visual inspection and some quantitative measures, and the second one is by analyzing the impact of the methodology on the accuracy of the segmentation using neural networks.

\subsection{Quality Assessment by Different Measurement}
The first way of assessing the outcome of the interpolation methods of the spectral pixels was done visually by 2D and 3D comparative plots and is described below.

\subsubsection{Plots}

The interpolation results for an image were compared both with the reference and among themselves. This was done by the generation of two distinct types of graphical representations: 2D plots of given pixels and 3D surface plots of the entire image. In this way, we can identify which interpolation result models better the shape of the original sample. Moreover, it provides early indications if certain interpolation types are unsuitable for a specific dataset. However, this information alone cannot validate or invalidate the interpolation because its goal is not only to preserve the initial information but also to augment the quantity of information contained.

\begin{figure}[htbp]
    \centering
    \includegraphics[width=0.7\linewidth]{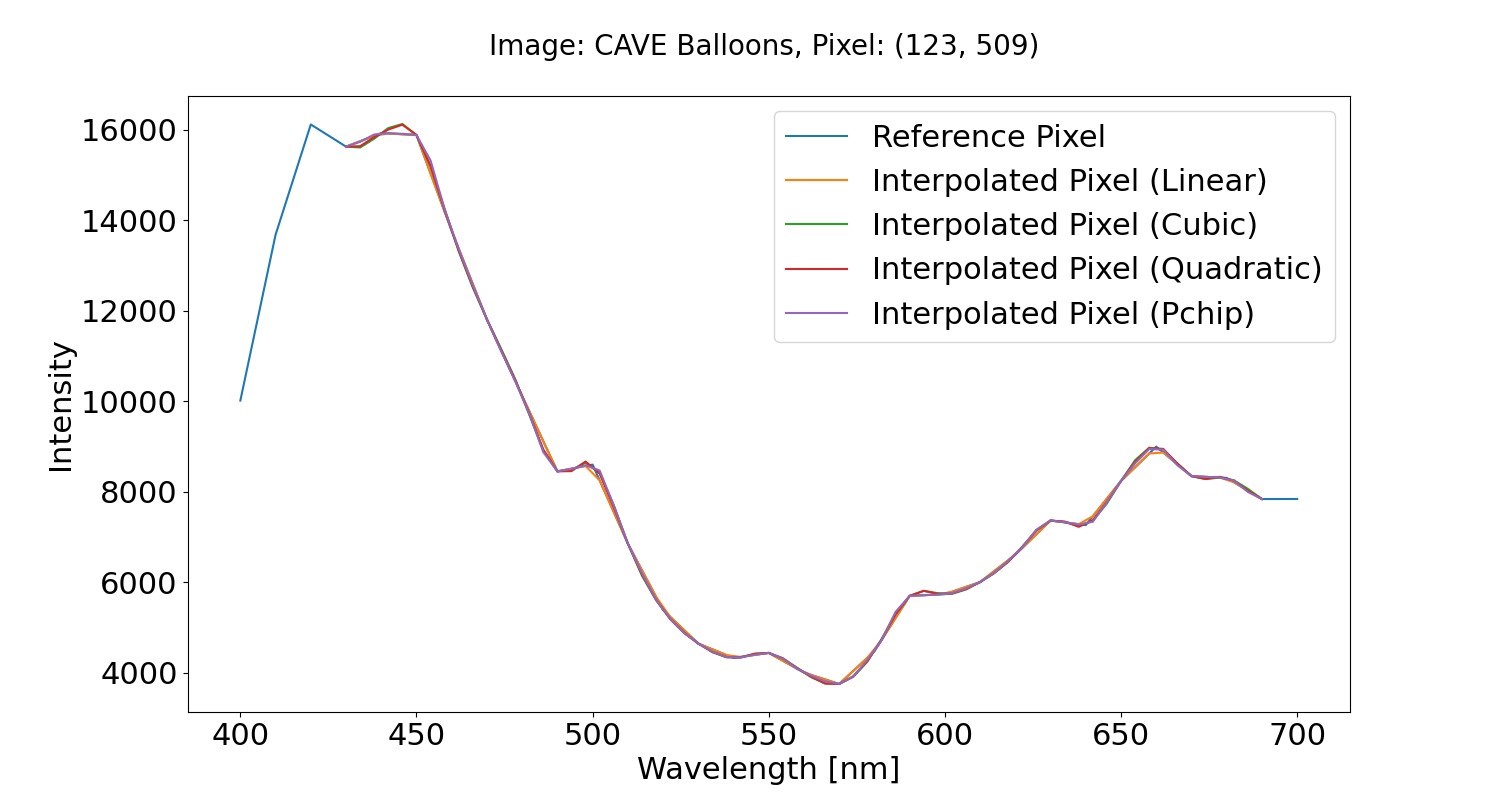}
    \caption{Reference and Interpolated Pixel for CAVE Balloons}
    \label{fig:2d_plot_cave}
\end{figure}

\begin{figure}[htbp]
    \centering
    \includegraphics[width=0.7\linewidth]{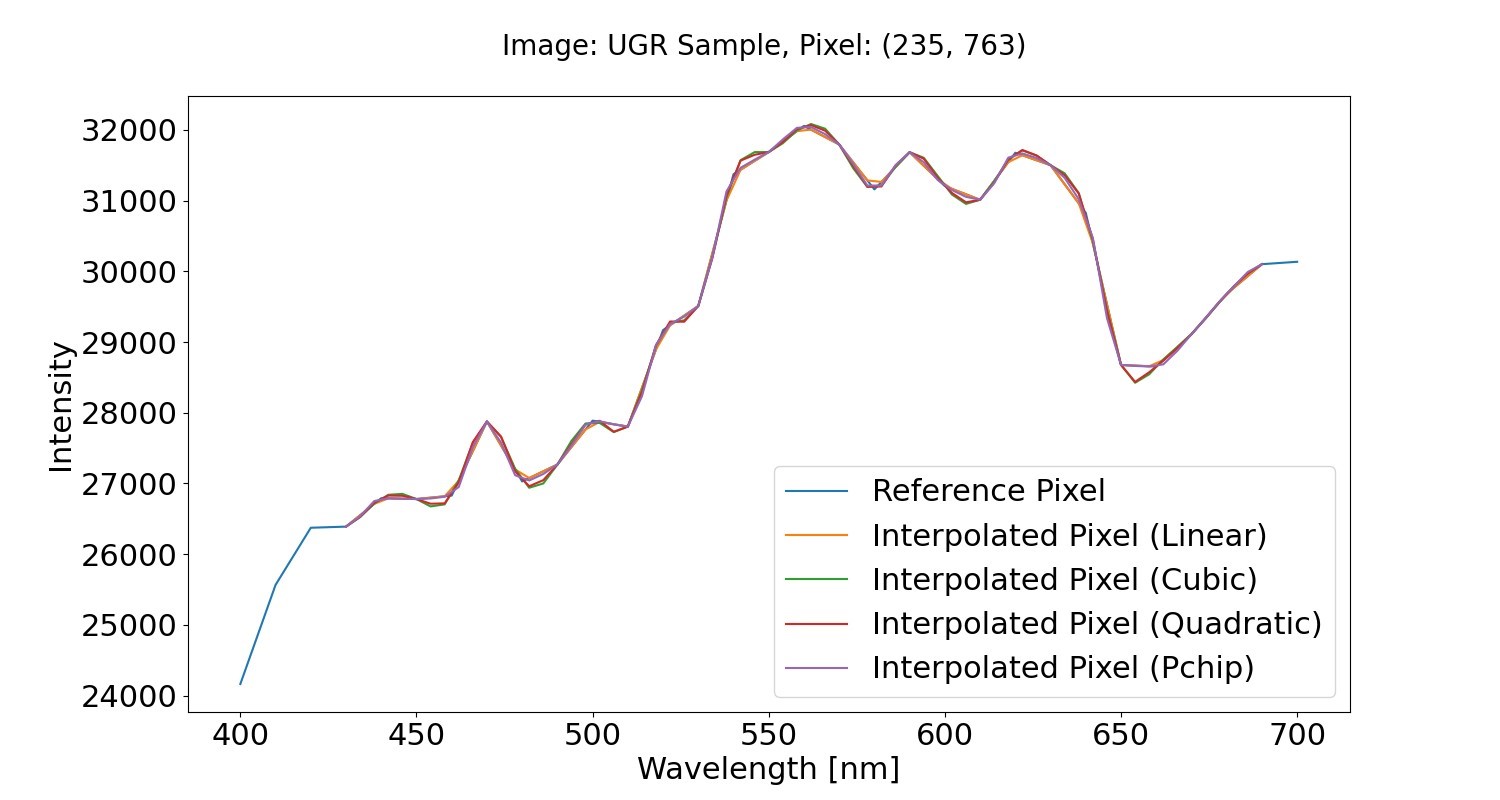}
    \caption{Reference and Interpolated Pixel for UGR}
    \label{fig:2d_plot_ugr}
\end{figure}

\begin{figure}[htbp]
    \centering
    \includegraphics[width=0.7\linewidth]{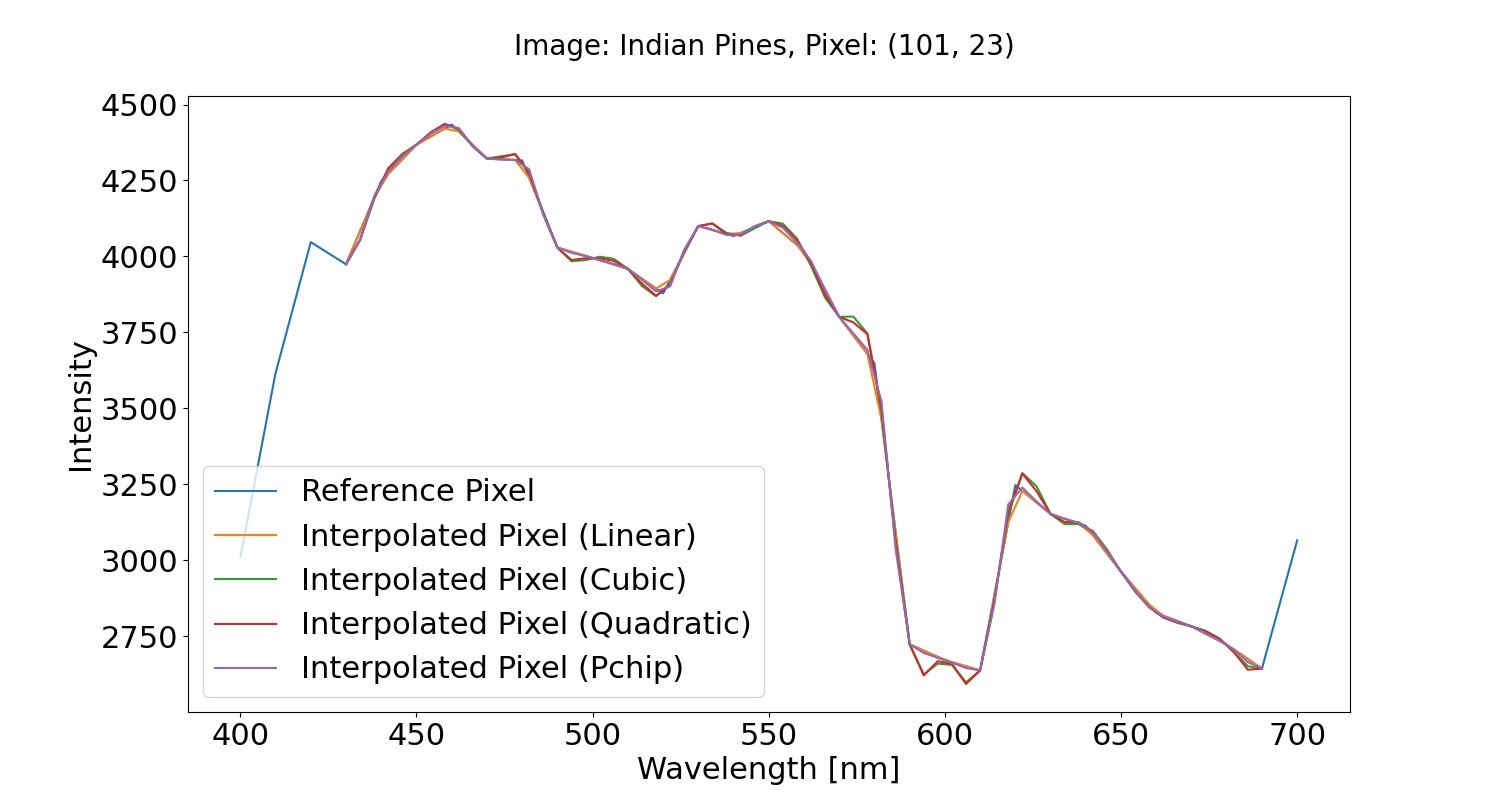}
    \caption{Reference and Interpolated Pixel for Indian Pines}
    \label{fig:2d_plot_indian_pines}
\end{figure}

\begin{figure}[htbp]
    \centering
    \includegraphics[width=0.7\linewidth]{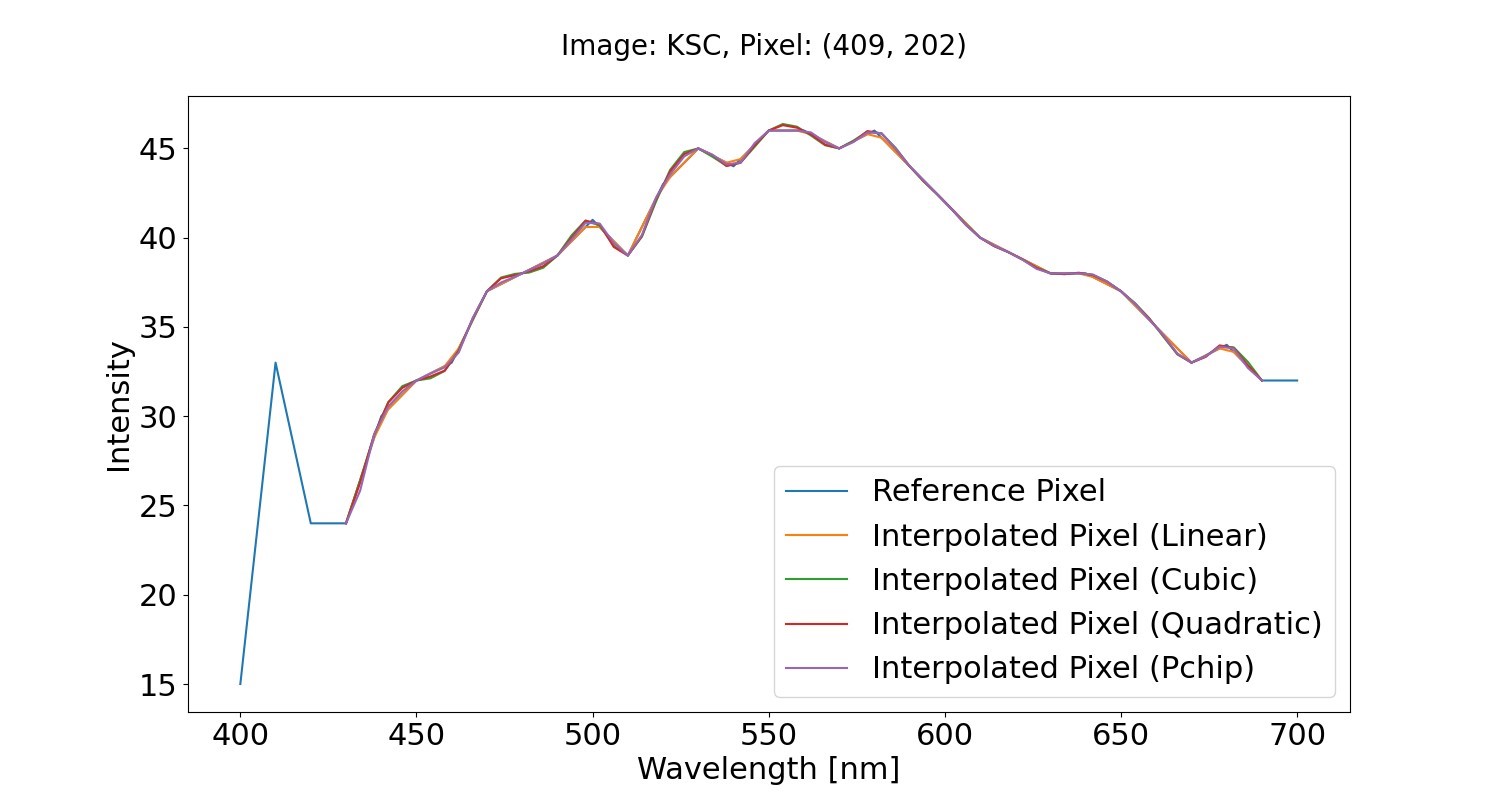}
    \caption{Reference and Interpolated Pixel for KSC}
    \label{fig:2d_plot_ksc}
\end{figure}

\begin{figure}[htbp]
    \centering
    \includegraphics[width=0.7\linewidth]{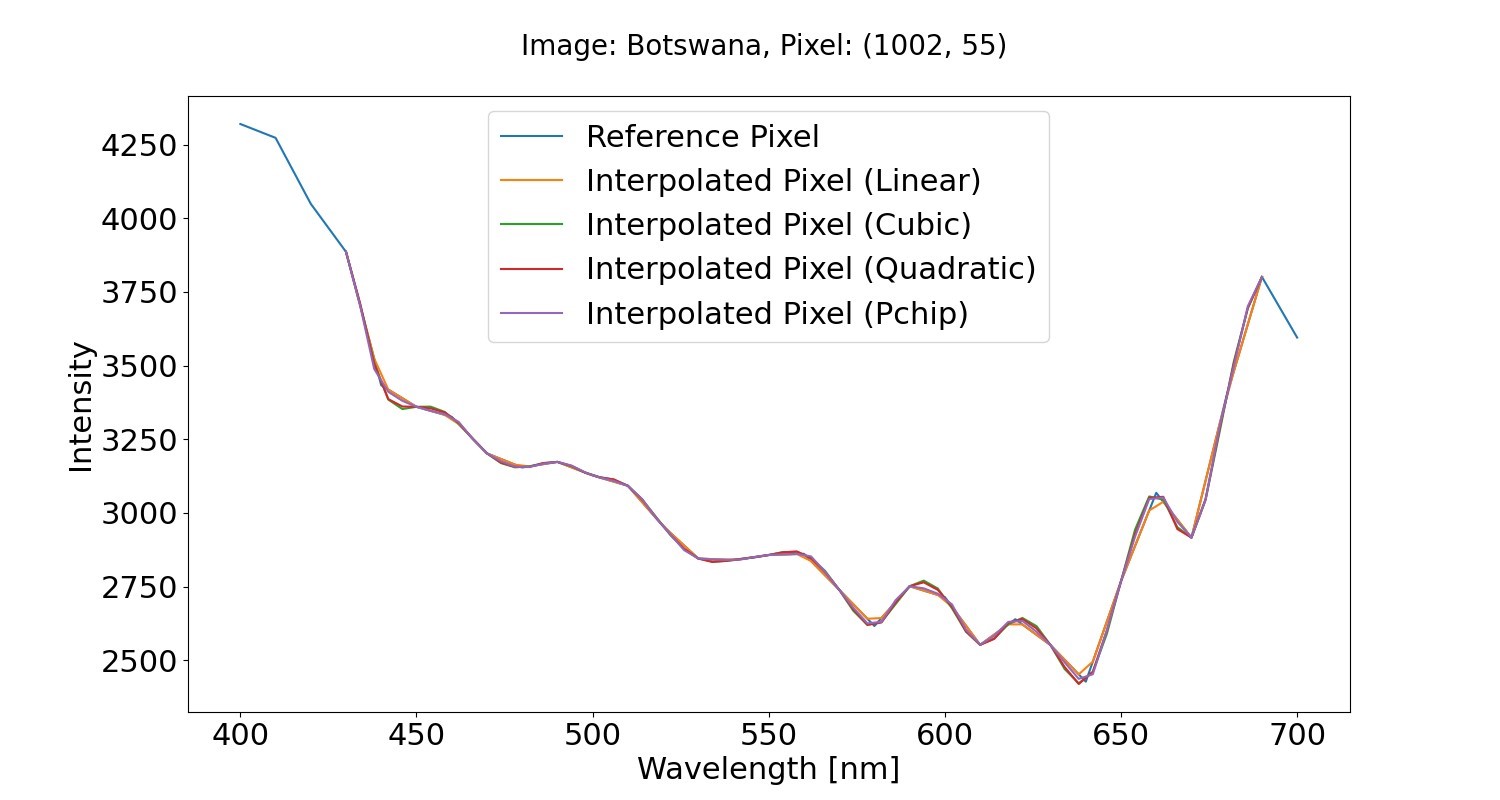}
    \caption{Reference and Interpolated Pixel for Botswana}
    \label{fig:2d_plot_botswana}
\end{figure}

For the 2D plots, we selected a random pixel and represented its spectral signature graphically. Each plot contains the pixel from the original image as well as all the interpolated versions. Results on pixels from each dataset are illustrated in Figures \ref{fig:2d_plot_cave} (CAVE), \ref{fig:2d_plot_ugr} (UGR), \ref{fig:2d_plot_indian_pines} (Indian Pines), \ref{fig:2d_plot_ksc} (KSC), and  \ref{fig:2d_plot_botswana} (Botswana). 

\begin{figure}[htbp]
    \centering
    \includegraphics[width=0.7\linewidth]{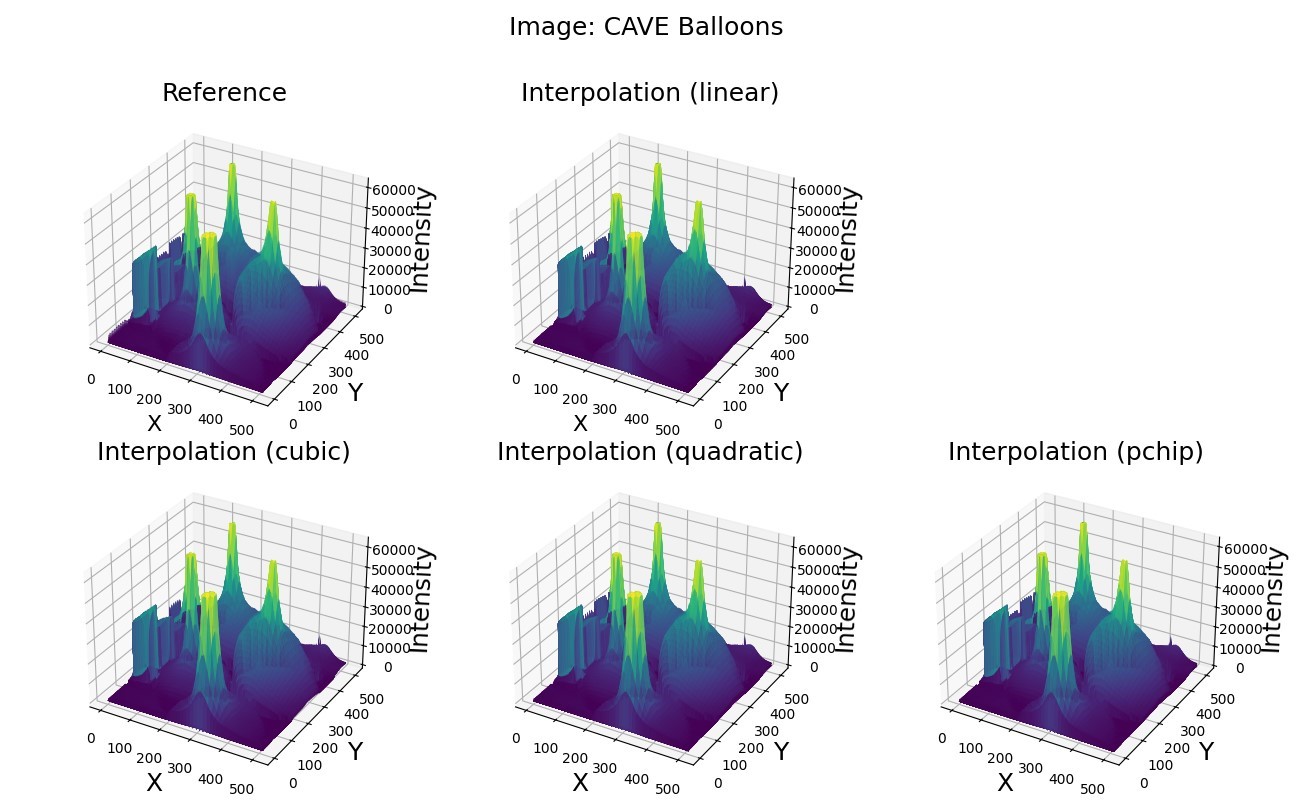}
    \caption{Pixels Surface for CAVE Balloons}
    \label{fig:3d_plot_cave}
\end{figure}

\begin{figure}[htbp]
    \centering
    \includegraphics[width=0.7\linewidth]{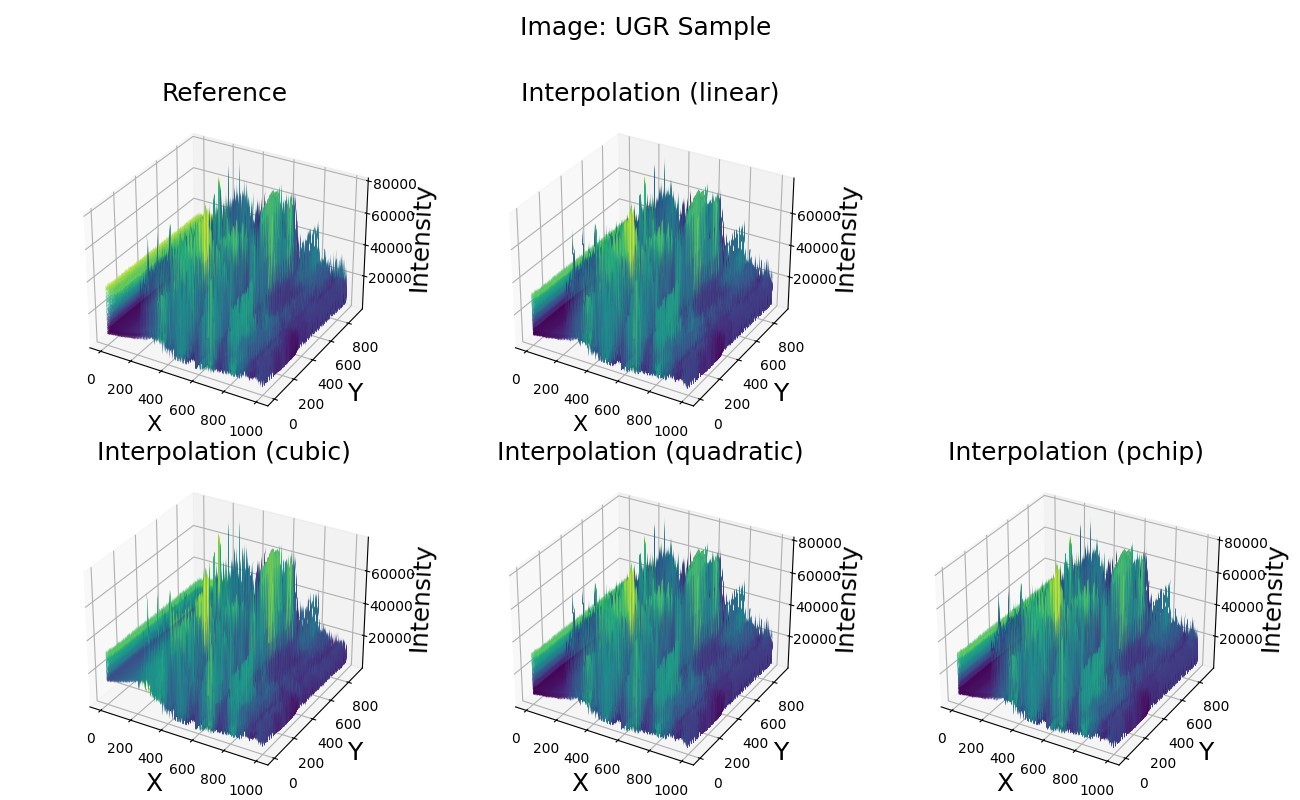}
    \caption{Pixels Surface for UGR}
    \label{fig:3d_plot_ugr}
\end{figure}

\begin{figure}[htbp]
    \centering
    \includegraphics[width=0.7\linewidth]{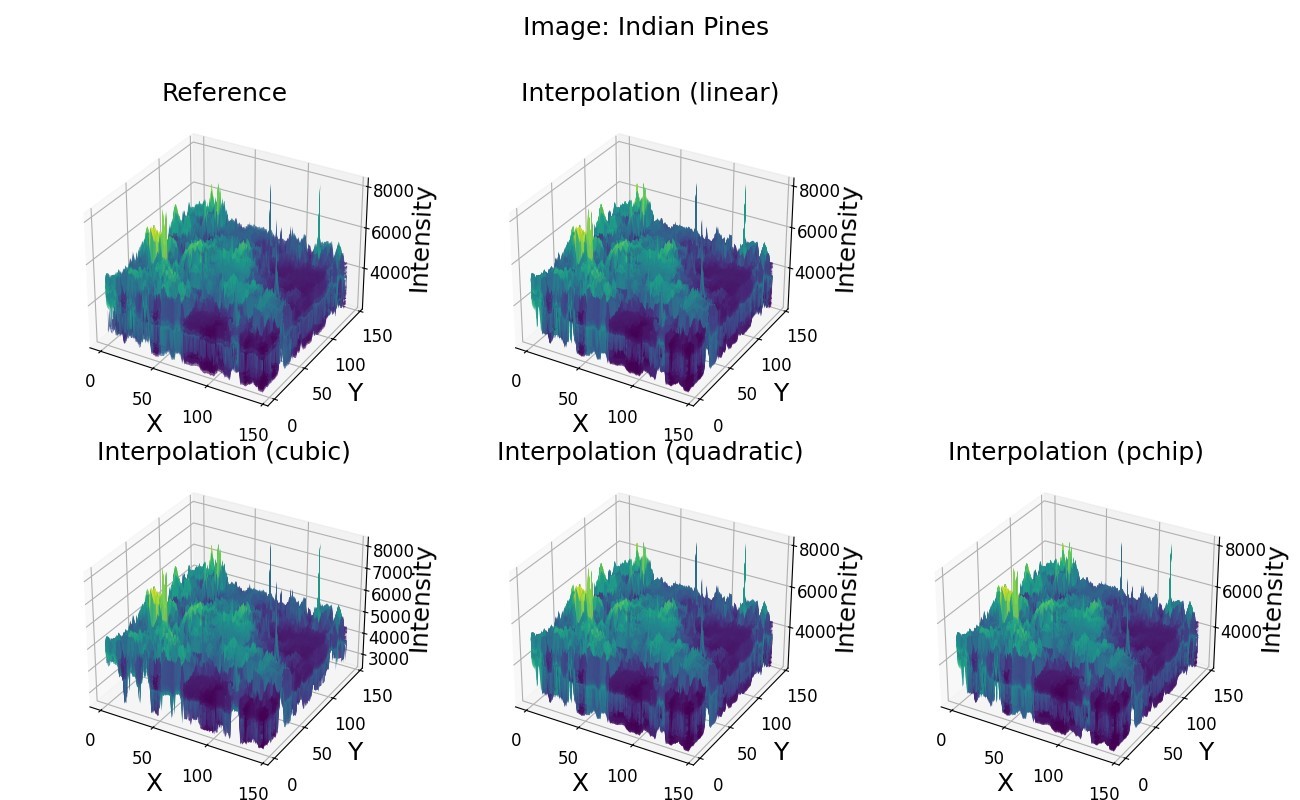}
    \caption{Pixels Surface for Indian Pines}
    \label{fig:3d_plot_indian_pines}
\end{figure}

\begin{figure}[htbp]
    \centering
    \includegraphics[width=0.7\linewidth]{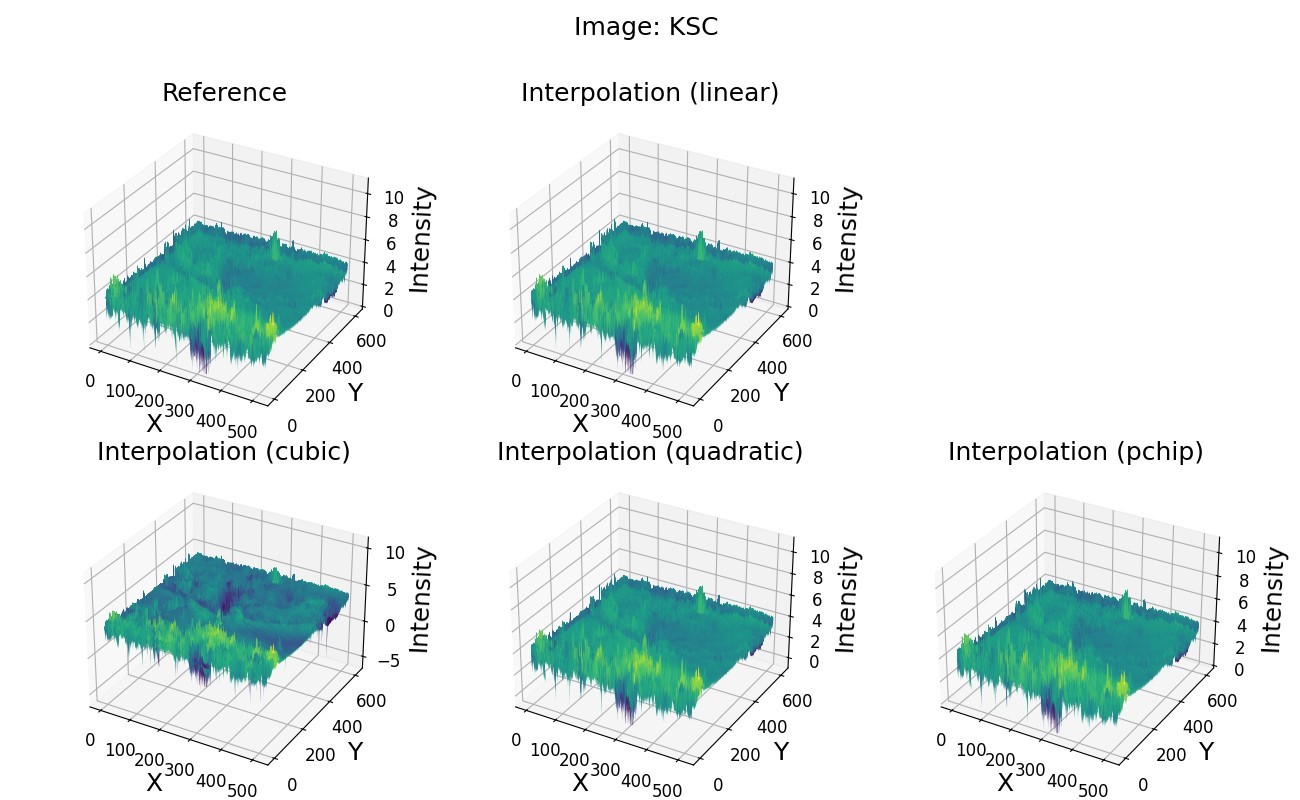}
    \caption{Pixels Surface for KSC}
    \label{fig:3d_plot_ksc}
\end{figure}

\begin{figure}[htbp]
    \centering
    \includegraphics[width=0.7\linewidth]{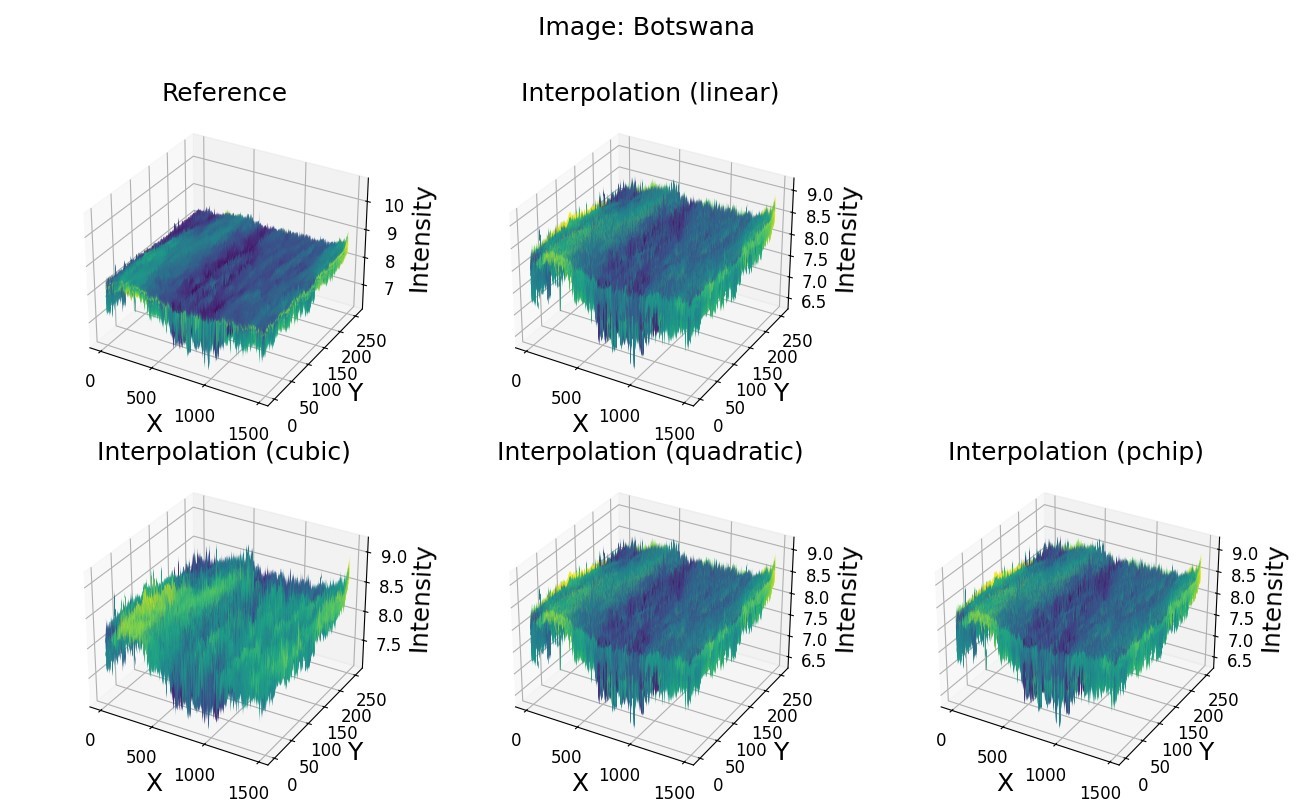}
    \caption{Pixels Surface for Botswana}
    \label{fig:3d_plot_botswana}
\end{figure}

The 3D plots offer a comprehensive visualization of the pixel surface distribution across the entire image. They also contain the reference dataset together with the results of the interpolation methods. Some plots can be seen in Figures \ref{fig:3d_plot_cave} (one image of CAVE dataset), \ref{fig:3d_plot_ugr} (one image of UGR dataset), \ref{fig:3d_plot_indian_pines} (Indian Pines), \ref{fig:3d_plot_ksc} (KSC), and \ref{fig:3d_plot_botswana} (Botswana).

By analyzing the graphical representations, we can affirm that all selected interpolation techniques provide satisfactory results. They predominantly follow the original shape of the pixel, but there are also different changes in intensities. The tests performed on the neural networks will provide a better perception, enabling us to estimate the impact of these alterations. 

\subsubsection{Surface Differences}

There are a lot of well-known metrics in the academic literature used for assessing the quality of MS and HS images. These include but are not limited to, Spectral Discrepancy, Correlation Coefficient, Mean Square Error (MSE), Root Mean Square Error (RMSE), Spectral Angle Mapper (SAM), Standard Deviation, Peak Signal to Noise Ratio (PSNR), and Image Fidelity \cite{super_resolution_q_a}, \cite{comparison_q_a}. Nevertheless, these metrics are extremely valuable only if the compared images have the same number of spectral bands. 

In the context of this study, we increased the number of spectral bands, starting with a spectral resolution of 10 nm and interpolating the intensities for the missing wavelengths to have a spectral resolution of 4 nm. Given these circumstances, we opted to use as one of the validation methods the results provided by computing the surfaces under the 2D plots using the Trapezoidal Rule \cite{trapezoidal_rule} and comparing the average differences between the reference and the interpolations. This approach offers a more precise comparison than the visual one offered by the plots, as it enables a quantitative assessment. This criterion allows us to identify if specific methods are not appropriate to be used on different datasets, as indicated by exceptionally high or low values compared to the rest of the results. Table \ref{tab:surfaces_under_plot} presents a summary of the obtained results for these surfaces.

The results for the interpolations across all datasets have a high degree of similarity, with no prominent outliers, suggesting that all proposed methods are valid. Nevertheless, depending on the dataset, we can choose the most appropriate interpolation technique depending on the task that we further want to solve. 

\begin{table}[htbp]
\centering
\tiny
\begin{tabular}{|c|l|r|}
\hline
Dataset & \multicolumn{1}{c|}{Interpolation} & \multicolumn{1}{c|}{Surface Average Difference} \\
\hline
\multirow{4}{*}{CAVE Balloons} & Linear & 3.60\\
\cline{2-3} & Cubic & 3.56\\ 
\cline{2-3} & Quadratic & 3.60\\
\cline{2-3} & PCHIP & 3.56\\ 
\hline
\multirow{4}{*}{UGR} & Linear & 3.90\\ 
\cline{2-3} & Cubic & 3.89\\
\cline{2-3} & Quadratic & 3.88\\
\cline{2-3} & PCHIP & 3.90\\
\hline
\multirow{4}{*}{Indian Pines} & Linear & 1.89\\
\cline{2-3} & Cubic & 1.88\\
\cline{2-3} & Quadratic & 1.95\\
\cline{2-3} & PCHIP & 2.13\\ 
\hline
\multirow{4}{*}{KSC} & Linear & 6.01\\
\cline{2-3} & Cubic & 7.32\\
\cline{2-3} & Quadratic & 6.53\\ 
\cline{2-3} & PCHIP & 5.54\\
\hline
\multirow{4}{*}{Botswana} & Linear & 5.90\\ 
\cline{2-3} & Cubic & 5.80\\
\cline{2-3} & Quadratic & 5.79\\ 
\cline{2-3} & PCHIP & 5.87\\
\hline
\end{tabular}
\caption{Surface Average Differences}
\label{tab:surfaces_under_plot}
\end{table}

\subsubsection{Custom Mean Squared Error (CMSE) Metric}
In order to consistently evaluate and compare the different interpolation methods, we searched for an appropriate metric. Except for the plot-based measures, we could not find in literature such a measure, and therefore, we tried to adapt one of the most used metrics for the evaluation of results, i.e., the Mean Squared Error (MSE). The MSE usually compares two signals/functions with the same data points, yielding the degree of similarity between them. In our case, it could compare two MS pixels that have the same spectral signature. Alternatively, MSE could be used in the case of interpolation if a ground truth exists to compare with, which, in our case, does not exist. 

In the case of interpolating MS images, we want to compare an interpolated pixel with the original one and evaluate how well the resulting interpolated MS pixel respects the shape of the original one; the MSE formula cannot be directly used.
If we assume that the interpolated pixel faithfully replicates the shape of the original one, then interpolating back on the source wavelengths should produce a result very similar to the source pixel. As a consequence, we elaborated a custom mean square error (CMSE) measure, which, for each MS pixel, calculates the MSE between this pixel and the result of its interpolation onto the desired wavelengths and then back onto the original ones. We could express this formula by:
$$CMSE(p) = \frac{(p-I_b(I_f(p))^2}{nw}$$
in which $I_f(p)$ represents the \textit{forward interpolation}, i. e the interpolation of pixel $p$ on the new wavelengths and $I_b(p)$ represents the \textit{backward} interpolation of $p$, i.e. the interpolation onto the source wavelength, and $nw$ the number of wavelengths.

\begin{figure}[htbp]
    \centering
    \includegraphics[width=\linewidth]{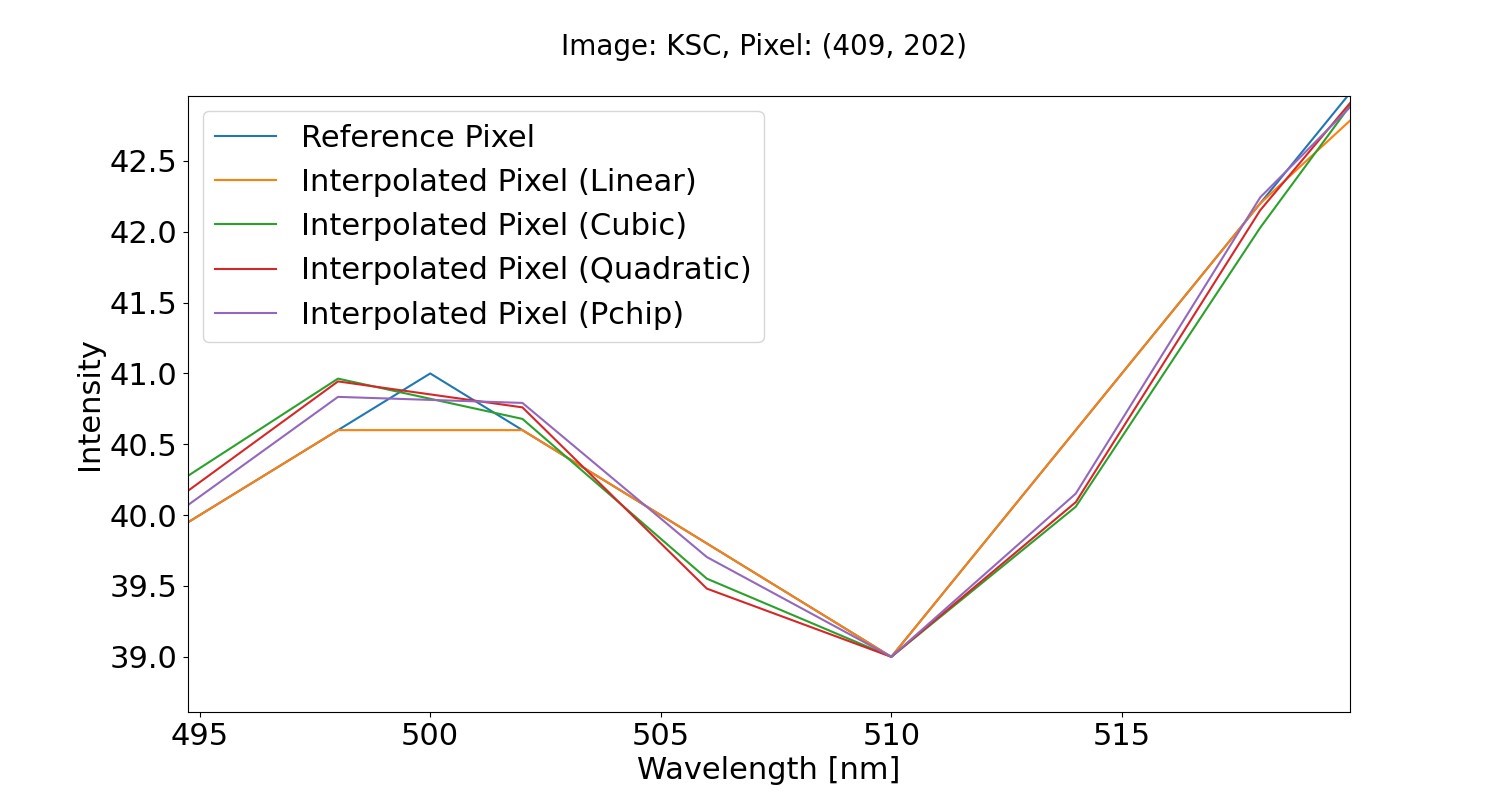}
    \caption{Magnified plot for a selected pixel from the KSC dataset.}
    \label{fig:KSC_zoomed_in}
\end{figure}

The results of this measure are represented in Table \ref{tab:mse_ref_interpolated_interpolation}. It can be observed that, for the linear interpolation, the CMSE is very large, while for the quadratic and cubic, it is relatively small.
In our opinion, this is an expected result, as linear interpolation is usually the most rough approximation for a signal. This observation underlines the results observed in the plot presented in Figure \ref{fig:KSC_zoomed_in}, which presents a magnified section from the 2D plot of a pixel of the KSC dataset.

\begin{table}[htbp]
\centering
\tiny
\begin{tabular}{|c|l|r|}
\hline
Dataset & \multicolumn{1}{c|}{Interpolation} & \multicolumn{1}{c|}{MSE} \\ 
\hline
\multirow{4}{*}{CAVE Balloons} & Linear & 3675.01\\
\cline{2-3} & Cubic & \textbf{7.75}\\
\cline{2-3} & Quadratic & \textbf{5.27}\\
\cline{2-3} & PCHIP & 341.31\\
\hline
\multirow{4}{*}{UGR} & Linear & 7800.99\\
\cline{2-3} & Cubic & \textbf{23.12}\\
\cline{2-3} & Quadratic & \textbf{15.08}\\
\cline{2-3} & PCHIP & 897.71\\
\hline
\multirow{4}{*}{Indian Pines} & Linear & 554.50\\
\cline{2-3} & Cubic & \textbf{1.65}\\
\cline{2-3} & Quadratic & \textbf{1.09}\\ 
\cline{2-3} & PCHIP & 78.59\\
\hline
\multirow{4}{*}{KSC} & Linear & 96.17\\
\cline{2-3} & Cubic & \textbf{0.58}\\
\cline{2-3} & Quadratic & \textbf{0.37}\\
\cline{2-3} & PCHIP & 8.76 \\
\hline
\multirow{4}{*}{Botswana} & Linear & 311.14\\
\cline{2-3} & Cubic & \textbf{0.84}\\ 
\cline{2-3} & Quadratic & \textbf{0.57}\\
\cline{2-3} & PCHIP & 35.75\\
\hline
\end{tabular}
\caption{CMSE: Reference - Interpolated Interpolation}
\label{tab:mse_ref_interpolated_interpolation}
\end{table}

\subsubsection{Normalized Difference Vegetation Index (NDVI)}
In the context of MS and HS imaging processing, NDVI is a pivotal indicator for precision agriculture, providing farmers with crucial data on crop health and resource allocation \cite{whispers2023}. The red (RED) and near-infrared (NIR) spectral bands of the MS image are used to calculate the NDVI, which is a vegetation index. The general formula is given by Equation \ref{eqn:eq1}.
\begin{equation}
\label{eqn:eq1}
    NDVI = \frac{NIR - RED}{NIR + RED}
\end{equation}

The NDVI values range within the interval $[-1,1]$. They can be interpreted as follows: 
\begin{itemize}
\item $(0, 0.33]$ for bare soil with little to no vegetation cover; 
\item $(0.33, 0.66]$ for unhealthy or sparse vegetation; 
\item $(0.66, 1]$ for dense and healthy vegetation; 
\item $[-1, 0]$ for water or inanimate things. 
\end{itemize}

Our objective was to investigate the influence of interpolation on the calculation of the NDVI. From the conducted tests, as shown in Table \ref{tab:mse_ndvi_ref_interpolation}, all the methods appear equally suitable. This finding is significant as it suggests that the most straightforward and quickest method, linear interpolation, can be effectively employed in this context.
\begin{table}[htbp]
\centering
\tiny
\begin{tabular}{|c|l|r|}
\hline
Dataset              & \multicolumn{1}{c|}{Interpolation} & \multicolumn{1}{c|}{MSE for NDVI} \\ \hline
\multirow{4}{*}{Indian Pines} & Linear                             & 0.00014                                                \\ \cline{2-3} 
                              & Cubic                              & 0.00024                                               \\ \cline{2-3} 
                              & Quadratic                          & 0.00020                                               \\ \cline{2-3} 
                              & PCHIP                              & 0.00019                                               \\ \hline

\multirow{4}{*}{KSC} & Linear                             & 0.00024                                               \\ \cline{2-3} 
                     & Cubic                              & 0.00042                                               \\ \cline{2-3} 
                     & Quadratic                          & 0.00038                                               \\ \cline{2-3} 
                     & PCHIP                              & 0.00038                                               \\ \hline

\multirow{4}{*}{Botswana} & Linear                             & 0.00075                                               \\ \cline{2-3} 
                          & Cubic                              & 0.00155                                               \\ \cline{2-3} 
                          & Quadratic                          & 0.00139                                               \\ \cline{2-3}  
                          & PCHIP                              & 0.00117                                              \\ \hline
\end{tabular}
\caption{MSE: NDVI Reference - NDVI Interpolation}
\label{tab:mse_ndvi_ref_interpolation}
\end{table}

\subsection{Quality Assessment by the Accuracy of the Semantic Segmentation}

In MS and HS imaging, semantic segmentation plays a crucial role by enabling pixel-wise labeling of the image. In this context increasing the number of samples of a dataset could be beneficial for the training of neural network models. This goal could be achieved by using interpolation for fusing several datasets, as proposed in section \ref{sec:materials_methods}. In this subsection, we evaluate and compare the proposed fusion method in the context of training neural networks for semantic segmentation of MS and HS images. 

For this purpose, two different neural networks were implemented. The first one is a simple Fully Connected Neural Network (FCNN), while the second architecture, UNet, a Convolutional Neural Network (CNN), is known for providing good accuracy for remote sensing image semantic segmentation tasks.

In practical applications, Fully Connected Neural Networks are used in a wide range of domains including object recognition \cite{fcnn_object_recognition}. On the other hand, UNet architectures provide robust solutions for a variety of tasks outperforming traditional methods and even human experts in certain cases including semantic segmentation tasks in remote sensing areas \cite{semantic_seg_remote_sensing1}, \cite{semantic_seg_remote_sensing2}, \cite{semantic_seg_remote_sensing3} and precision agriculture tasks \cite{conv_net_agriculture}, among many others \cite{unet_medicine}, \cite{unet_tumor_segmentation},\cite{unet_data_reconstruction}. 


\subsubsection{FCNN}

\begin{figure}[htbp]
    \centering
    \includegraphics[width=\linewidth]{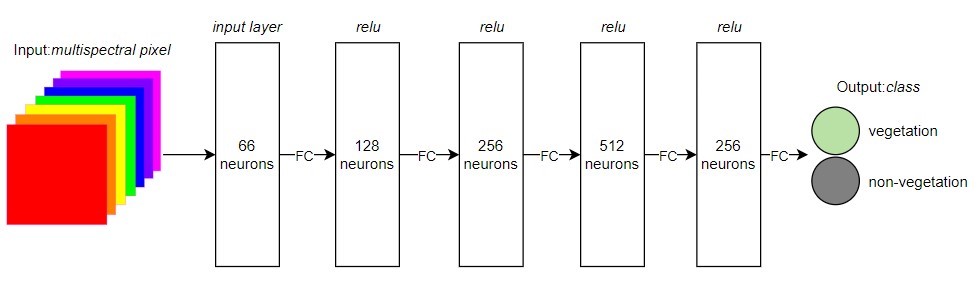}
    \caption{Network Model}
    \label{fig:network_model}
\end{figure}

Assuming that the input data was preprocessed and the number of wavelengths present in all the images that will be used is 66, the input layer of the neural network will correspondingly contain 66 neurons. The network architecture includes four fully connected hidden layers and an output layer with two neurons for the two classification categories: vegetation and non-vegetation. To leverage the computational efficiency associated with powers of 2, the hidden layers contain 128, 256, 512, and 256 neurons, respectively. The Rectified Linear Unit (ReLU) is the activation function used for the hidden layers, while the Softmax function is used for the output layer to convert the output into probabilities. Figure \ref{fig:network_model} provides a schematic representation of the model architecture.

The network was trained on the Pavia University image and various other dataset combinations that are detailed in Section \ref{subsec:net_results}. The pixels were combined into a single set, which was then shuffled and divided into training and testing subsets. After several tests, it was determined that 150 epochs were sufficient for the classification task. We used the Cross-Entropy Loss as a loss function, and we selected Adam as the optimizer, with a learning rate of 0.0001.
After shuffling all the pixels, the dataset was split into training (80\%) and testing (20\%) subsets and random batches of 2048 pixels were used for training.  


\subsubsection{UNet}

An UNet model is characterized by its U-shape, which contains a contracting path, or encoder, to capture the context and a symmetric expanding path, or decoder, for precise localization \cite{unet}. In the implemented architecture, both the encoder and decoder consist of five 3D convolutional layers, followed by a Batch Normalization layer and a ReLU activation function. The last layer in the decoder path is a convolutional layer with the number of filters equal to the number of classes, which is two, and it uses a Softmax activation function. The output of the encoder's last layer and the decoder's last layer are concatenated, facilitating better localization because the high-resolution features from the encoder are combined with the upsampled output of the decoder. We used 3D convolutional layers and image patches of 10 x 10 pixels with 66 channels for each pixel, setting the input shape to (10, 10, 66, 1). The filters were selected based on the computational benefits of operations with powers of two. Therefore, for the contracting path, they were set to 64, 64, 128, 256, and 256 respectively. We used 64, 128, 256, 512, and 256 for the expanding path, respectively.

The network was trained on $10 \times 10$ patches from Pavia University and various other combinations of patches from different datasets that are presented in detail in Subsection \ref{subsec:net_results}. The patches combined from all the datasets were split into training, validation, and testing subsets. Depending on the selected datasets, the number of epochs varies because the training is stopped when a monitored metric, the loss in our case, has stopped improving. As an optimizer, we use Adam with a learning rate of 0.001.

To expand the dataset used in the training process, we used the data augmentation technique \cite{augmentation}. For each patch in the training dataset, we generated three more images using a horizontal flip, a vertical flip, and a rotation of the patch with a random angle between -180 and 180 degrees.


\subsubsection{Training Datasets}
\label{subsec:net_results}

In order to train and test the results on these networks, four of the exposed datasets can be used, namely: Pavia University, KSC, Botswana, and Indian Pines, because they also contain a ground truth image with the associated labels. These labels were merged into only two main classes: Vegetation and Non-Vegetation. 

We tested the neural networks on different training scenarios using both unprocessed datasets and combinations of original and interpolated sets. Table \ref{tab:fcnn_results} presents the most representative results obtained for the semantic segmentation using a fully connected neural network, while Table \ref{tab:unet_results} displays a part of the results for the UNet architecture.

\begin{table}[htbp]
\centering
\tiny
\begin{tabular}{|c|c|l|r|}
\hline
{\begin{tabular}[c]{@{}c@{}}Training\\ Dataset \end{tabular}} & {\begin{tabular}[c]{@{}c@{}}Training\\ Accuracy \end{tabular}}        & \multicolumn{1}{c|}{\begin{tabular}[l]{@{}c@{}}Testing Dataset\\ (Interpolation) \end{tabular}} & \multicolumn{1}{c|}{\begin{tabular}[c]{@{}c@{}}Testing\\ Accuracy \end{tabular}}\\ \hline
\multirow{12}{*}{Pavia University} & \multirow{12}{*}{71.5\%} & KSC (Linear)& 72.56  \\
\cline{3-4} &  & KSC (Cubic)  & 72.37 \\
\cline{3-4} &  & KSC (Quadratic) & 72.37\\ 
\cline{3-4}&  & KSC (PCHIP)  & 72.56 \\
\cline{3-4} &  & Botswana (Linear) & 53.17 \\
\cline{3-4} &   & Botswana (Cubic) & 53.48\\
\cline{3-4} &  & Botswana (Quadratic) & 53.17\\ \cline{3-4} &  & Botswana (PCHIP) & 53.17\\
\cline{3-4} &  & Indian Pines (Linear) & 90.75 \\ \cline{3-4} &  & Indian Pines (Cubic) & 90.62\\
\cline{3-4} &  & Indian Pines (Quadratic) & 90.61\\ \cline{3-4} &  & Indian Pines (PCHIP)  & 90.64 \\
\hline
\multirow{12}{*}{\begin{tabular}[c]{@{}c@{}}Pavia University\\ and\\ Botswana (Linear)\end{tabular}} & \multirow{12}{*}{88\%}  & KSC (Linear)  & 72.56\\ \cline{3-4} & & KSC (Cubic)& 72.40  \\
\cline{3-4} &   & KSC (Quadratic) & 72.54 \\
\cline{3-4} & & KSC (PCHIP) & 72.56 \\
\cline{3-4}  &   & Botswana (Linear)& 90.02 \\
\cline{3-4} & & Botswana (Cubic)  & 89.93\\
\cline{3-4}  &  & Botswana (Quadratic)& 89.93\\
\cline{3-4} &  & Botswana (PCHIP)& 89.99\\
\cline{3-4} & & Indian Pines (Linear)& 94.98\\
\cline{3-4} & & Indian Pines (Cubic)& 94.84\\
\cline{3-4} & & Indian Pines (Quadratic)& 94.89\\
\cline{3-4} &  & Indian Pines (PCHIP) & 94.97\\
\hline
\multirow{12}{*}{\begin{tabular}[c]{@{}c@{}}Pavia University\\ and\\ KSC (Cubic)\end{tabular}} & \multirow{12}{*}{89.27\%}   & KSC (Linear) & 97.08\\
\cline{3-4} & & KSC (Cubic) & 97.12\\
\cline{3-4} & & KSC (Quadratic) & 97.12\\
\cline{3-4} &  & KSC (PCHIP)& 97.08\\
\cline{3-4} & & Botswana (Linear)& 77.46\\
\cline{3-4} &  & Botswana (Cubic)& 77.77 \\
\cline{3-4} & & Botswana (Quadratic)& 77.83\\
\cline{3-4} &  & Botswana (PCHIP)& 78.14\\
\cline{3-4} & & Indian Pines (Linear)& 92.97 \\
\cline{3-4} &  & Indian Pines (Cubic)& 92.46 \\
\cline{3-4} & & Indian Pines (Quadratic) & 92.54\\
\cline{3-4} & & Indian Pines (PCHIP)& 92.87\\
\hline
\end{tabular}
\caption{FCNN Training and Testing Results}
\label{tab:fcnn_results}
\end{table}

\begin{table}[htbp]
\centering
\tiny
\begin{tabular}{|c|c|l|r|}
\hline
{\begin{tabular}[c]{@{}c@{}}Training\\ Dataset \end{tabular}} & {\begin{tabular}[c]{@{}c@{}}Training\\ Accuracy \end{tabular}}        & \multicolumn{1}{c|}{\begin{tabular}[c]{@{}c@{}}Testing Dataset\\ (Interpolation) \end{tabular}} & \multicolumn{1}{c|}{\begin{tabular}[c]{@{}c@{}}Testing\\ Accuracy \end{tabular}}\\ \hline
\multirow{12}{*}{\begin{tabular}[c]{@{}c@{}}Pavia University\\ (82 epochs)\end{tabular}} & \multirow{12}{*}{89\%} & KSC (Linear) & \multirow{4}{*}{72.08}\\
\cline{3-3} & & KSC (Cubic) & \\
\cline{3-3} & & KSC (Quadratic)& \\
\cline{3-3} & & KSC (PCHIP)& \\
\cline{3-4} & & Botswana (Linear)& 91.70\\
\cline{3-4} & & Botswana (Cubic) & 88.98\\
\cline{3-4} & & Botswana (Quadratic)& 90.52\\
\cline{3-4} & & Botswana (PCHIP) & 90.36\\
\cline{3-4} & & Indian Pines (Linear)& 69.41\\
\cline{3-4} & & Indian Pines (Cubic) & 71.19\\
\cline{3-4} & & Indian Pines (Quadratic)& 71.44\\
\cline{3-4} & & Indian Pines (PCHIP)& 71.13\\
\hline
\multirow{12}{*}{\begin{tabular}[c]{@{}c@{}}Pavia University\\ and\\ Indian Pines (Linear)\\ (95 epochs)\end{tabular}} & \multirow{12}{*}{89\%} 
 & KSC (Linear) & \multirow{4}{*}{72.08}\\
 \cline{3-3} & & KSC (Cubic)& \\
 \cline{3-3} & & KSC (Quadratic) & \\
 \cline{3-3} & & KSC (PCHIP)& \\
 \cline{3-4} & & Botswana (Linear)& 86.11\\
 \cline{3-4} & & Botswana (Cubic) & 86.10\\
 \cline{3-4} & & Botswana (Quadratic) & 86.11\\
 \cline{3-4} & & Botswana (PCHIP) & 86.11\\
 \cline{3-4} & & Indian Pines (Linear) & 94.96\\
 \cline{3-4} & & Indian Pines (Cubic) & 94.95\\
 \cline{3-4} & & Indian Pines (Quadratic) & 94.96\\
 \cline{3-4} & & Indian Pines (PCHIP) & 94.96\\
 \hline
\multirow{12}{*}{\begin{tabular}[c]{@{}c@{}}Pavia University\\ and\\ KSC (Cubic)\\ (89 epochs)\end{tabular}} & \multirow{12}{*}{86\%} & KSC (Linear) & \multirow{4}{*}{72.08} \\
\cline{3-3} & & KSC (Cubic) & \\
\cline{3-3} & & KSC (Quadratic)& \\
\cline{3-3} & & KSC (PCHIP) & \\
\cline{3-4} & & Botswana (Linear)& 92.43\\
\cline{3-4} & & Botswana (Cubic)& 92.44\\
\cline{3-4} & & Botswana (Quadratic) & 92.43\\
\cline{3-4} & & Botswana (PCHIP)& 92.43\\
\cline{3-4} & & Indian Pines (Linear) & 80.04 \\
\cline{3-4} & & Indian Pines (Cubic) & 78.79\\
\cline{3-4} & & Indian Pines (Quadratic) & 78.70\\
\cline{3-4} & & Indian Pines (PCHIP) & 78.83\\
\hline
\end{tabular}
\caption{UNet Training and Testing Results}
\label{tab:unet_results}
\end{table}

We have also experimented with evaluating the FCNN using the whole Pavia University image, including all 103 spectral bands. This choice was chosen in order to take advantage of the extra data found in the supplementary spectral bands, which may be quite valuable in the case of spectral imaging. As a result, we carried out the interpolations this time, taking into account the wavelengths between 430 and 838 nm.

At first, the outcomes were less than encouraging, indicating a training accuracy of only 33\%. This underperformance can be attributed to the specific architecture chosen for the neural network. Considering that we've nearly doubled the number of spectral bands, and consequently increased the neurons in the input layer, we recognized the potential benefit of also doubling the neurons in certain hidden layers. Our final tests were done on an architecture that includes four hidden layers with 256, 512, 256, and 128 neurons, respectively. In this setup, we observed noticeable improvements during the testing phase. We trained the neural network on the Pavia University dataset for 150 epochs and then tested the classification on the KSC, Botswana, and Indian Pines datasets. The results are presented in Table \ref{tab:fcnn_results_103_bands}. We noticed significant progress with the Botswana dataset, where the accuracy nearly doubled compared to the previous results. For KSC, the results were roughly the same, while for Indian Pines they improved by 5\%, resulting in a total of 95\% correct classifications. 

\begin{table}[htbp]
\centering
\tiny
\begin{tabular}{|c|c|l|r|}
\hline
{\begin{tabular}[c]{@{}c@{}}Training\\ Dataset \end{tabular}} & {\begin{tabular}[c]{@{}c@{}}Training\\ Accuracy \end{tabular}}        & \multicolumn{1}{c|}{\begin{tabular}[l]{@{}c@{}}Testing Dataset\\ (Interpolation) \end{tabular}} & \multicolumn{1}{c|}{\begin{tabular}[c]{@{}c@{}}Testing\\ Accuracy \end{tabular}}\\ \hline
\multirow{12}{*}{Pavia University} & \multirow{12}{*}{66.26\%} & KSC (Linear) & \multirow{4}{*}{72.56} \\
\cline{3-3} & & KSC (Cubic) &\\
\cline{3-3} & & KSC (Quadratic) &\\
\cline{3-3} & & KSC (PCHIP) &\\
\cline{3-4} & & Botswana (Linear) & \multirow{4}{*}{82.85}\\
\cline{3-3} & & Botswana (Cubic) & \\
\cline{3-3} & & Botswana (Quadratic) & \\
\cline{3-3} & & Botswana (PCHIP) & \\
\cline{3-4} & & Indian Pines (Linear)& \multirow{4}{*}{95.33}\\
\cline{3-3} & & Indian Pines (Cubic) & \\
\cline{3-3} & & Indian Pines (Quadratic) & \\
\cline{3-3} & & Indian Pines (PCHIP) & \\ \hline
\end{tabular}
\caption{FCNN Training and Testing Results using 103 spectral bands}
\label{tab:fcnn_results_103_bands}
\end{table}

Given the diversity across the datasets, incorporating a wide array of both vegetation and non-vegetation associated pixels, we regard the outcomes as promising. Notably, we observed improvements in the outcomes when the training incorporated a mix of unprocessed pixels from the Pavia University dataset and results from the interpolations. Moreover, in all the cases, the results from the testing phase are better than the training ones for at least one of the three tested datasets. For the others, the accuracy is similar to the training phase's. The lower results are caused by the differences in the dataset labels. Sometimes, the vegetation is associated with the color green, while in other cases, it is correlated with dry vegetation revealing shades of brown. In such scenarios, we have observed that combining pixels from two different images significantly improves the quality of the results.

\section{Discussions and Conclusions}
\label{discussions_conclusions}

This paper illustrates a preprocessing approach for the MS and HS data fusion task. After performing the interpolation of the spectral pixels with respect to a chosen reference wavelength spectrum and validating the results using three distinct methods, we concluded that the results obtained are encouraging and demonstrate the validity of the proposed solution. 

This study was developed on six different datasets. However, the selection of these datasets posed a serious challenge because of the variety of characteristics of the spectral images and the labels provided for them, making it hard to decide if they can or cannot be fused. This still remains a problem and represents a potential area for further enhancement of our methodology. Each MS or HS dataset has a specific description that can provide useful information and insights for gathering a collection of sets that can be fused. Therefore, developing an algorithm that analyzes these descriptions and determines whether the data can be aggregated would be the next step in improving our solution. This would also allow us to test and interpret the results on larger datasets. 

Nonetheless, the exposed results based on the four manually selected datasets, which were used in the context of validation using neural networks specialized in semantic segmentation, showed to be promising, even though the merged labels could not be as accurate as desired and the segmentation networks were relatively simple ones. Therefore, the proposed fusion technique proves to be efficient, robust, and adaptable, showing potential to be used in different applications and to gain in accuracy, when provided with better labels or used with more sophisticated architectures.

\section*{Acknowledgments}
This work was funded by European Union. The AI4AGRI project entitled “Romanian Excellence Center on Artificial Intelligence on Earth Observation Data for Agriculture” received funding from the European Union’s Horizon Europe research and innovation program under grant agreement no. 101079136.


\bibliographystyle{elsarticle-num-names} 
\bibliography{newRef.bib}





\end{document}